\documentclass{article}

\usepackage{microtype}
\usepackage{graphicx}
\usepackage{subfigure}
\usepackage{booktabs} % for professional tables

\usepackage{times}
\usepackage{epsfig}
\usepackage{amsmath}
\usepackage{amssymb}

\graphicspath{ {./images/} }
\usepackage{float}
\usepackage{sidecap}
\usepackage{booktabs}

\usepackage{hyperref}

\usepackage[accepted]{icml2020}

\icmltitlerunning{Image-to-Image Translation with Text Guidance}

\begin{document}

\twocolumn[
\icmltitle{Image-to-Image Translation with Text Guidance}



\icmlsetsymbol{equal}{*}

\begin{icmlauthorlist}
\icmlauthor{Bowen Li,}{cs}
\icmlauthor{Xiaojuan Qi,}{cs}
\icmlauthor{Philip H. S. Torr,}{cs}
\icmlauthor{Thomas Lukasiewicz}{cs} \\
\end{icmlauthorlist}

\centering
\text{University of Oxford} \\
\text{\{bowen.li, thomas.lukasiewicz\}@cs.ox.ac.uk} \\ \text{\{xiaojuan.qi, philip.torr\}@eng.ox.ac.uk}
%\icmlaffiliation{cs}{University of Oxford}

\icmlkeywords{Machine Learning, ICML}

\vskip 0.3in
]




\begin{abstract}
The goal of this paper is to embed controllable factors, i.e., natural language descriptions, into image-to-image translation with generative adversarial networks, which allows text descriptions to determine the visual attributes of synthetic images. We propose four key components: (1) the implementation of part-of-speech tagging to filter out non-semantic words in the given description, (2) the adoption of an affine combination module to effectively fuse different modality text and image features, (3) a novel refined multi-stage architecture to strengthen the differential ability of discriminators and the rectification ability of generators, and (4) a new structure loss to further improve discriminators to better distinguish real and synthetic images. Extensive experiments on the COCO dataset demonstrate that our method has a superior performance on both visual realism and semantic consistency with given descriptions.
%the superior performance of the proposed method. 
\end{abstract}

\section{Introduction}
\label{sec:intro}
\begin{figure*}[t]
\begin{minipage}{1\textwidth}
\begin{minipage}{0.161\textwidth}
\centering
\small{Segmentation mask.}
\end{minipage}
\begin{minipage}{0.161\textwidth}
\centering
\small{A \textbf{stop sign} is in a \textbf{grassy} rural area.}
\end{minipage}
\noindent\begin{minipage}{0.161\textwidth}
\centering
\small{A \textbf{pizza} with \textbf{cheese} and \textbf{pepperoni} is on a wooden tray.}
\end{minipage}
\;\;\noindent\begin{minipage}{0.161\textwidth}
\centering
\small{Segmentation mask.}
\end{minipage}
\noindent\begin{minipage}{0.161\textwidth}
\centering
\small{A \textbf{green} glass vase holding several stems of \textbf{green} flowers.}
\end{minipage}
\noindent\begin{minipage}{0.161\textwidth}
\centering
\small{A \textbf{purple} glass vase holding several stems of \textbf{purple} flowers.}
\end{minipage}
\end{minipage}

\begin{minipage}{1\textwidth}
\begin{minipage}{0.161\textwidth}
\includegraphics[width=1\linewidth, height=1\linewidth]{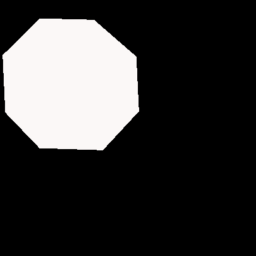}
\end{minipage}
\begin{minipage}{0.161\textwidth}
\includegraphics[width=1\linewidth, height=1\linewidth]{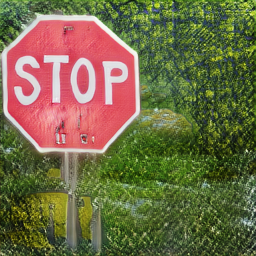}
\end{minipage}
\noindent\begin{minipage}{0.161\textwidth}
\includegraphics[width=1\linewidth, height=1\linewidth]{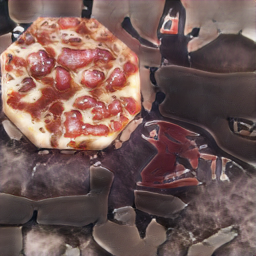}
\end{minipage}
\;\;\begin{minipage}{0.161\textwidth}
\includegraphics[width=1\linewidth, height=1\linewidth]{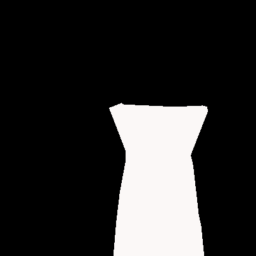}
\end{minipage}
\noindent\begin{minipage}{0.161\textwidth}
\includegraphics[width=1\linewidth, height=1\linewidth]{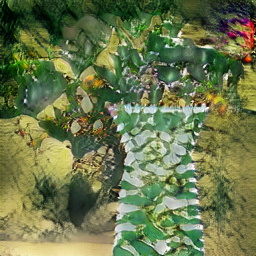}
\end{minipage}
\noindent\begin{minipage}{0.161\textwidth}
\includegraphics[width=1\linewidth, height=1\linewidth]{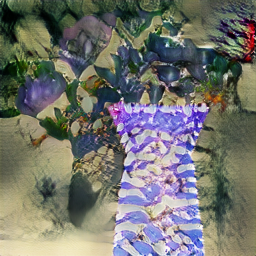}
\end{minipage}
\end{minipage}
\smallskip

\begin{minipage}{1\textwidth}
\begin{minipage}{0,161\textwidth}
\centering
\small{Segmentation mask.}
\end{minipage}
\begin{minipage}{0,161\textwidth}
\centering
\small{A \textbf{zebra} in a \textbf{road} with \textbf{trees} in the background.}
\end{minipage}
\begin{minipage}{0,161\textwidth}
\centering
\small{A \textbf{giraffe} is standing on a \textbf{grass} covered field.}
\end{minipage}
\;\;\noindent\begin{minipage}{0,161\textwidth}
\centering
\small{Segmentation mask.}
\end{minipage}
\noindent\begin{minipage}{0,161\textwidth}
\centering
\small{A large \textbf{white} passenger jet flying through a \textbf{blue} sky.}
\end{minipage}
\noindent\begin{minipage}{0,161\textwidth}
\centering
\small{A large \textbf{yellow} passenger jet flying through a \textbf{grey} sky.}
\end{minipage}
\end{minipage}

\begin{minipage}{1\textwidth}
\begin{minipage}{0,161\textwidth}
\includegraphics[width=1\linewidth, height=1\linewidth]{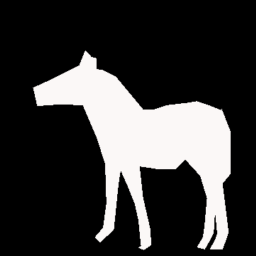}
\end{minipage}
\begin{minipage}{0,161\textwidth}
\includegraphics[width=1\linewidth, height=1\linewidth]{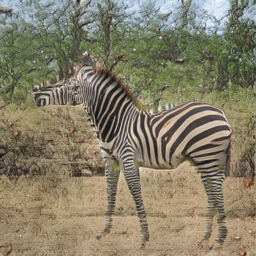}
\end{minipage}
\begin{minipage}{0,161\textwidth}
\includegraphics[width=1\linewidth, height=1\linewidth]{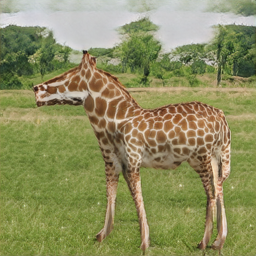}
\end{minipage}
\;\;\noindent\begin{minipage}{0,161\textwidth}
\includegraphics[width=1\linewidth, height=1\linewidth]{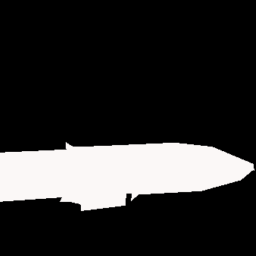}
\end{minipage}
\noindent\begin{minipage}{0,161\textwidth}
\includegraphics[width=1\linewidth, height=1\linewidth]{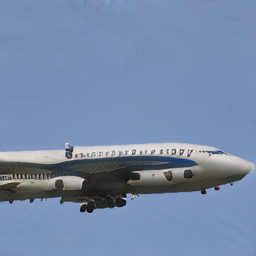}
\end{minipage}
\noindent\begin{minipage}{0,161\textwidth}
\includegraphics[width=1\linewidth, height=1\linewidth]{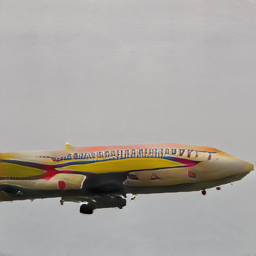}
\end{minipage}
\end{minipage}

\centering
\caption{Given a segmentation mask and a text provided by a user that describes desired objects and visual attributes, the goal of this model is to generate realistic images semantically matching the given descriptions with the global structure defined by the masks.}
\label{fig:intro}
\end{figure*}

Conditional image synthesis aims to generate realistic images semantically matching given conditions, including text-to-image generation \cite{reed2016generative, zhang2017stackgan, zhang2018stackgan++, xu2018attngan, li2019controllable} and image generation from scene graphs \cite{johnson2018image, ashual2019specifying}, semantic layout \cite{isola2017image, chen2017photographic, wang2018high, mo2018instagan, park2019semantic}, or coarse layout \cite{zhao2019image}, which enables numerous potential applications in many areas, including design, video games, art, architecture, and image editing. 

The goal of this paper is to produce realistic images from segmentation masks, and also to embed controllable factors, i.e., natural language descriptions, into the generation process to control the visual attributes (e.g., colour, background, and texture) of synthetic images semantically matching the given texts. Unlike current state-of-the-art image-to-image translation models \cite{isola2017image, chen2017photographic, wang2018high,  park2019semantic, pavllo2019controlling, tang2019local}, which require fine-grained pixel-labelled semantic maps to decide what to generate and usually fail to predict exact visual attributes of synthetic images, our model is able to generate desired images under the control of natural language descriptions, even if the provided masks are simple. As shown in Fig.~\ref{fig:intro}, given a simple circle segmentation mask, our model is able to generate a \emph{stop sign at grassy area} and also a \emph{pizza with cheese and pepperoni}.
%and rectangle binary segmentation masks, our model is able to generate a donut, an orange, a apple and a pizza matching the circle mask, and to produce different colour buses semantically aligned with given text descriptions.

%but unlike those only focusing on generation process \cite{park2019semantic, wang2018high, chen2017photographic, pavllo2019controlling},
%we also aims to incorporate controllable factors, e.g., texts, to control the visual attributes (e.g., colour, background, texture) of synthetic images which semantically match given texts. As shown in Fig.XX, the synthetic images of current models are unpredictable and mainly depends on the training dataset. In other words, given a animal like segmentation, how do we know it is a zebra, giraffe, cow, sheep or horse? Thus, what kinds of visual attributes should be generated by current methods? Indeed, the outlook mainly is decided by training datasets.

%In order to address above problem, our paper focuses on incorporating controllable factors, i.e., natural language, into generation process, such that users can modify the segmentation to control global structure of synthetic images, e.g., the position, size, pose, type and number of objects, and decide visual attributes of output according to texts, in one framework.

To achieve this, the key is to completely disentangle different visual attributes contained in text descriptions and images, and then to build an accurate correlation between semantic words and corresponding visual attributes to achieve effective control. Also, how to effectively generate realistic images involving different modality representations (e.g., natural language) on more difficult datasets (e.g., COCO \cite{lin2014microsoft}) is a critical issue to address, where each image in the dataset has multiple objects with complicated relationships between each other.

%when the generation involves different modality information (e.g., text and image) in more complicated datasets, e.g., COCO \cite{lin2014microsoft}, the models usually fail to produce realistic images. One of the main reason is few training examples in the COCO dataset discussed in \cite{li2019controllable}. Therefore, the other key point is to effectively solve this few-shot problem along with dataset. %in particular, when an image has multiple target instances and a translation task involves significant changes in shape,

To address the above issues, we propose a novel generative adversarial network, called RefinedGAN, which can effectively generate realistic images from segmentation masks, and also embed controllable factors (i.e., text descriptions) in the generation process, which allows users to determine the exact visual attributes of synthetic images. 

We propose four key components in our RefinedGAN: (1) part-of-speech (POS) tagging is implemented to filter out less important (non-semantic) words in the given text description, (2) the affine combination module  \cite{li2019manigan} is adopted to effectively fuse different modality representations (text and segmentation mask), and to also build an effective connection between them, (3) to generate high-quality images from segmentation masks involving text, we propose a novel refined multi-stage architecture for discriminators and generators, which strengthens the differential ability of discriminators and in turn encourages generators at lower stages to produce not only the global structure and layout, but also fine-grained details as much as possible. Also, this architecture improves the rectification ability of generators to complete missing details and correct inappropriate attributes produced from lower stages,
%in order to strengthen their generation and fake image distinction abilities, respectively. 
%used to distinguish part of images produced from higher stages in order to strengthen their fake image distinction ability. 
and (4) a new structure loss is proposed to further improve discriminators in order to better distinguish fake images from real ones. 
%As shown in the Fig.xx, our full model can generate high-quality images and also allow users to effective control the global structure using segmentation map, and modify visual attributes using natural language descriptions.

Finally, an extensive analysis is performed, which demonstrates that our method can effectively generate realistic images on the more complex COCO dataset \cite{lin2014microsoft}, and also accurately control the visual attributes of synthetic images using natural language descriptions. Experimental results on the dataset show that our method outperforms existing methods both qualitatively and quantitatively. 

%disentangle different visual attributes and accurately control the generation of the synthetic image using natural language descriptions. Also, experimental results on the COCO dataset \cite{lin2014microsoft} show that our method outperforms existing state of the art both qualitatively and quantitatively.

\section{Related Work}

\textbf{Image-to-image translation} is closely related to our work. \citeauthor{chen2017photographic}~\yrcite{chen2017photographic} achieved high-quality image generation using a single feedforward network. \citeauthor{wang2018high}~\yrcite{wang2018high} proposed multi-scale generator and discriminator architectures in order to generate high-resolution images. \citeauthor{mo2018instagan}~\yrcite{mo2018instagan} made use of object segmentation masks to achieve instance transfiguration. \citeauthor{park2019semantic}~\yrcite{park2019semantic} implemented affine transformation in conditional normalisation techniques to avoid information loss. However, all these works and others \cite{isola2017image, qi2018semi, tang2019local} only focus on generating realistic images from pixel-labelled semantic maps without the ability to determine the visual attributes of synthetic images.

\textbf{Text-to-image generation} has made great progress with the development of GANs \cite{goodfellow2014generative}, including image generation from text \cite{reed2016generative, zhang2017stackgan, zhang2018stackgan++, xu2018attngan, li2019controllable} and scene graphs \cite{johnson2018image, ashual2019specifying}. Also, \citeauthor{hong2018inferring}~\yrcite{hong2018inferring} and \citeauthor{li2019object}~\yrcite{li2019object} proposed to generate intermediate layout first (i.e., bounding boxes and segmentation masks) and then convert the semantic layout into an image. 
However, all the above methods mainly focus on generating realistic images from text descriptions or scene graphs, and usually fail to produce high-quality results on more complicated datasets, such as COCO.

\textbf{Text-guided image manipulation} is about editing given images using texts to achieve semantic consistency. \citeauthor{dong2017semantic}~\yrcite{dong2017semantic} built an encoder-decoder architecture to get an appropriate modification. \citeauthor{nam2018text}~\yrcite{nam2018text} proposed a text-adaptive discriminator to utilise word-level information. Instead of using the simple and coarse concatenation method, \citeauthor{li2019manigan}~\yrcite{li2019manigan} proposed a novel affine combination module to effectively fuse different modality representations. However, these methods mainly aim to edit a given image rather than producing new results.

\textbf{Multi-stage architectures} have been widely adopted in GAN-based models \cite{denton2015deep, huang2017stacked,zhang2017stackgan, zhang2018stackgan++, xu2018attngan, shaham2019singan, li2019controllable} to produce high-resolution images, which have a generator and a discriminator at each level of an image pyramid, and generate image progressively from coarse-to-fine scale. Differently from them, our model fully makes use of features produced at higher stages by feeding these features to discriminators at lower ones to improve their differential ability, which further benefit generators to improve their rectification ability as well. Our architecture is more suitable for realistic image generation with finer details and under control of natural language descriptions.
%Inspired by their success, we propose a novel refined multi-stage architecture suitable for realistic image generation with finer details and under control of natural language descriptions. %and also mitigate the negative impact due to few training examples.
\begin{figure*}[t]
\begin{minipage}{1\textwidth}
\centering
\includegraphics[width=1\linewidth, height=0.373\linewidth]{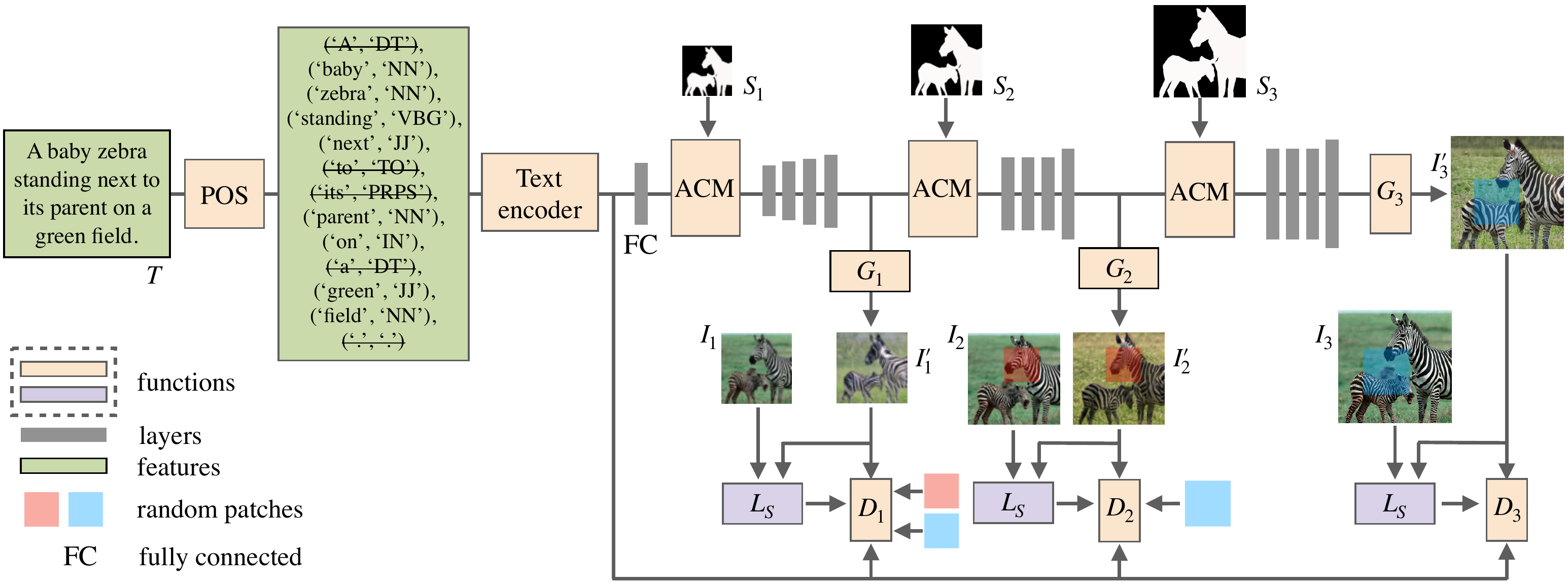}
\end{minipage}
\caption{The architecture of RefinedGAN. POS denotes the part-of-speech tagging. ACM denotes the affine combination module. $L_{S}$ denotes the structure loss defined in Sec.~\ref{sec:structure}. The attention is omitted for simplicity. Please see supplementary material for the full architecture. Note that in our paper, we just provide the original text descriptions instead of filtered words in each example for simplicity.}
\label{fig:archi}
\end{figure*}
\section{Refined Generative Adversarial Networks}
Suppose as given a segmentation mask $S$ and a text description $T$ provided by a user. The model aims to generate a realistic image $I'$ semantically matching the given text $T$ with the global structure defined by the segmentation mask $S$. To achieve this, we propose four components: (1) the implementation of part-of-speech (POS) tagging, (2) the adoption of an affine combination module \cite{li2019manigan}, (3) a refined multi-stage architecture for both discriminators and generators, and (4) a novel structure loss.

\subsection{Architecture}
We adopt the muti-stage ControlGAN \cite{li2019controllable} as our basic framework shown in Fig.~\ref{fig:archi}. We implement part-of-speech (POS) tagging to filter out non-semantic words in the given text description, and then feed the output into a pre-trained RNN \cite{xu2018attngan} to generate text representations. Then, we adopt an affine combination module (ACM) at each stage to fuse text features (generated from the previous stage) with the segmentation mask, which can build an accurate correlation between words and the corresponding semantic parts of the mask, and thus embed text information into the generation process enabling an effective controllable ability. 
Next, the fused features are refined by a residual block followed by an upsampling block to produce hidden features, which are fed into a generator to output synthetic images and also serve as the input for the next stage to produce images at a higher resolution. The whole framework generates high-quality images progressively, matching the global structure defined by the segmentation mask, and gradually produces regional visual attributes semantically aligned with the given description.

\subsection{Part-of-Speech Tagging}
\label{sec:pos}
Given a text description, it may contain some less important words that cannot help image generation and even cause negative impact. For example, as shown in Fig.~\ref{fig:archi}, words ``a, to, its" in the description do not have any semantic meaning, but if we keep these words, they may be connected with some visual attributes in the synthetic image, which may harm the ability of accurate control. Therefore, in order to filter out these words, we implement part-of-speech (POS) tagging to label each word such that each word in the description is marked up corresponding to a particular part of speech, based on both its definition and context, i.e., its relationship with adjacent and related words in the sentence \cite{bird2009natural}.

As shown in Fig.~\ref{fig:archi}, POS takes the text description as input and then labels each word with corresponding tags. In our model, we only keep words with specific tags: 
\begin{equation}
\text{NN}^{*}\textrm{, } \text{IN}^{*}\textrm{, } \text{VB}^{*}\textrm{, } \text{and JJ}^{*}
\textrm{,}
\end{equation}
where asterisk $*$ indicates zero or more occurrences of any element, $\text{NN}^{*}$ represents all nouns in different forms, $\text{IN}^{*}$ represents preposition or subordinating conjunction, $\text{VB}^{*}$ represents all verbs in any form, and $\text{JJ}^{*}$ represents all adjective. We only keep these specific words because nouns, prepositions and verbs already capture the main meaning of a sentence, and adjectives contain the major descriptions of visual attributes of an image.

\textbf{Why does filter out less important words work better than keeping all words?} First, not all words are equally important in a given text description, and some words may have no or even negative impact on the generation process, e.g., determiner (a, an, the), or adverb. Also, keeping these words, the generated word and sentence features from a RNN-based text encoder must contain these less useful information, and then an inappropriate correlation is built between non-semantic words and visual attributes. Thus, the performance of control may be affected, 
especially when users only want to modify some visual attributes of a synthetic image while preserving the other content. More details are discussed in Sec.~\ref{sec:ablation}.
%especially when users want to keep some already generated visual attributes and only modify some of them. More details are discussed in Sec.~\ref{sec:ablation}.   
%and then inappropriate correlation between non-semantic words and visual attributes is built by the model, which may affect the performance of control, especially when users want to keep some already generated visual attributes and only modify some of them. More details are discussed in Sec.~\ref{sec:ablation}.

%If we keep all words in the given text descriptions, first the generated sentence feature from the pre-trained RNN have these information and also treat all words equally, which  

% marking up a word in a text (corpus) as corresponding to a particular part of speech,[1] based on both its definition and its context—i.e., its relationship with adjacent and related words in a phrase, sentence, or paragraph. 

\subsection{Affine Combination Module}
To effectively fuse different modality features (i.e., text and image) together, we adopt the affine combination module (ACM) \cite{li2019manigan} instead of simply concatenating both features along the channel direction. In our model, the ACM is placed before each residual block followed by an upsampling block at the end of each stage, shown in Fig.~\ref{fig:archi}, and is defined as:
\begin{equation}
{h}'=h \odot W(S) + b(S),
\end{equation}
where $W$ and $b$ represent the functions that convert the segmentation mask $S$ to learned weights $W(S)$ and biases $b(S)$, $h$ denotes the hidden features encoded from the input text description, or it is the intermediate hidden representation between two stages, $h'$ denotes the fused features containing pieces of information of both language and segmentation mask, %two different modality information (i.e., language and segmentation mask), 
and $\odot$ denotes the Hadamard element-wise product. Please see the supplementary material for more details on the architecture.
%where $W(I)$ and $b(I)$ are the learned weights and biases based on the segmentation mask $I$, and $\odot$ denotes Hadamard element-wise product. We use $W$ and $b$ to represent the functions that convert the segmentation mask $I$ to scaling and bias values.

%placed each upsampling block at the end of each stage, shown in Fig.XX. 

%relate image regions with corresponding semantic words for generating new visual attributes semantically aligned with the given text description. Meanwhile, it also encodes original image representations for reconstructing text-irrelevant contents.

\subsection{Refined Multi-Stage Architecture}
%reuse existing discriminator instead of increase the number of stage with more generator and discriminator to produce high-quality images.
Generating realistic images involving different modality representations (e.g., natural language) on more difficult datasets (e.g., COCO) is a big challenge for generative models \cite{reed2016generative, dong2017semantic, nam2018text}, even with a multi-stage architecture \cite{zhang2018stackgan++, xu2018attngan, li2019controllable}, which generates a coarse image at the first stage, and then progressively increases its resolution with finer details.
%with multiple stages \cite{zhang2017stackgan, zhang2018stackgan++, xu2018attngan, li2019controllable}, which aim to produce a coarse image at the low stage, and then progressively increases the resolution of image with more details.
The ineffective generation is mainly because: (1) these models fail to produce a complete basic structure at lower stages, especially at the first one, which means some parts of the synthetic image generated at the first stage are unrealistic, and (2) generators lack the ability to complete missing details or rectify inappropriate visual attributes. Thus, due to the flawed basic image and less efficient generators, the models fail to generate high-quality images with realistic details everywhere.

In order to address the above issues, we propose a novel refined multi-stage architecture for both discriminators and generators, which can fully explore the internal distribution of patches within a single image to strengthen the differential ability of discriminators at lower stages and the rectification ability of generators.

As shown in Fig.~\ref{fig:archi}, our model has a multi-stage architecture and each stage has a generator and a discriminator, $\{G_{1}, D_{1}; G_{2}, D_{2}, ...\}$. %Generators are responsible for producing realistic images from coarse-to-fine scale, $\{I'_{1}, I'_{2}, ...\}$, 
Different-scale images are generated progressively, $\{I'_{1}, I'_{2}, ...\}$, and the resolution of the synthetic image is $4$ times of the previous one. The generation of an image starts at the coarsest scale with the smallest resolution and sequentially passes through higher stages to the finer scale with larger resolution. %Therefore, at the first stage, the generator produces the general structure of the image and the global layout of objects, and generators at the higher stages add finer details that are missed by previous stages.

To generate a complete structure at lower stages with finer details and thus to provide a better basic image for the following stages, we feed patches of real and fake images at higher stages to discriminators at lower ones, %As discussed in \cite{shaham2019singan}, 
where the internal distribution of patches within images at higher stages contains unseen but finer pieces of information, %to guide the model to produce realistic images. These finer pieces of information at higher stages 
which can be used as extra information to help to train and refine discriminators at lower stages to improve their differential ability, which in turn encourages generators at the same stages to produce a complete basic structure with fine-grained details, especially for the generator at the first stage, see Fig.~\ref{fig:refined}. Thus, the extra unconditional adversarial loss $\mathcal{L}_{Z_{Di}}$ for the discriminator at stage $i$ is defined as:
\begin{equation}
\begin{split}
\mathcal{L}_{Z_{Di}}=-(\sum_{k=i+1}^{K}(&E_{I_{k}\sim P_\text{data}}\left [ \log(D_{i}(P_{k})) \right ]+\\
&E_{{I}'_{k}\sim PG_{k}}\left [ \log(1-D_{i}({P}'_{k})) \right ]))
\textrm{,}
\end{split}
\label{eq:discriminator}
\end{equation}
where $K$ is the total number of stages, ${P}'_{k}$ and $P_{k}$ are random patches (detached) of the synthetic image $I'_{k}$ and the real image $I_{k}$ at a higher stage $k$, respectively. The size of patches ${P}'_{k}$ and $P_{k}$ matches the input requirement of the discriminator $D_{i}$.
%, and $P_{k}$ is a random patch of the real image $I_{k}$ at the higher stage $k$.
%facilitating to produce fine-grained images at first stage with global architecture and layout. 
%Then, based on this finer-scale image, following generators are able to generate much better high-quality results.

Besides, we further feed the informative patches of fake images produced at higher stages to these refined discriminators at lower ones in order to strengthen the rectification ability of their following generators, which can complete missing details and correct inappropriate visual attributes, shown in Fig.~\ref{fig:rectification}. Thus, the extra unconditional adversarial loss $\mathcal{L}_{Z_{Gi}}$ for the generator at stage $i$ is defined as:
\begin{equation}
\begin{split}
\mathcal{L}_{Z_{Gi}}=-(\sum_{k=1}^{i-1}{E_{{I}'_{i}\sim PG_{i}}\left [ \log(D_{k}({P}'_{k})) \right ]})
\textrm{,}
\end{split}
\label{eq:generator}
\end{equation}
where $i > 1$, ${P}'_{k}$ is a random patch (non-detached) of the $i^{th}$ stage synthetic image ${I}'_{i}$, and the cropped size of ${P}'_{k}$ matches the input requirement of the discriminator $D_{k}$.

\textbf{Why does the refined multi-stage architecture work better?} This architecture aims to refine discriminators and generators by feeding patches from higher stages to lower ones, where these patches contain an unseen internal distribution with fine-grained details. % for those discriminators and generators at low stages. 
By doing this, discriminators at lower stages can better identify fake images based on not only the global distribution, but also regional features, which in turn encourages generators at the same stages to produce realistic images with finer details. Also, these enhanced discriminators can provide regional feedback to generators at their following stages, refining the generators to produce realistic regions and have the abilities of completing missing contents and rectifying inappropriate visual attributes.

%the visual features produced at low stages can have much at limited size 

%A natural question may arise: the feature scales at low stages are much smaller than those at higher stage, so how to include more information? 

\subsection{Structure Loss}
\label{sec:structure}
To further improve the differential ability of discriminators, we propose a novel structure loss, which can also be used to stabilise the training, since generators have to produce natural statistics for both objects and background. More specifically, we use the provided segmentation mask to separate objects and background on both synthetic and real images. Then, we create new compositions with different objects and background, i.e., fake objects + real background and real objects + fake background, and feed these new images to discriminators to improve their differential ability, identifying a fake image if there exist some unrealistic regions only in the foreground or background. In turn, generators can be encouraged to produce finer details everywhere without preference, instead of focusing only on the generation of realistic objects or the background. 
%differential ability of discriminator to identify fake images when even part of it is still fake. In turn, the generator can be encouraged to produce more finer details at all regions instead of part of a image. 
Thus, the structure loss $\mathcal{L}_{S_{i}}$ at stage $i$ is defined as:
\begin{equation}
\begin{split}
\mathcal{L}_{S_{i}}=-E_{(X_{i}^{1},X_{i}^{2})}\left [ \log(D_{i}(X_{i}^{1})) \right ]+\left [ \log(D_{i}(X_{i}^{2})) \right ]
\textrm{,}
\end{split}
\label{eq:SL}
\end{equation}
where $X_{i}^{1}$ represents the new image composed of fake objects with real background, and $X_{i}^{2}$ denotes real objects with fake background at stage $i$.

%We improve the GAN loss in Eq. (2) by incorporating a feature matching loss based on the discriminator. This loss stabilizes the training as the generator has to produce natural statistics at multiple scales. Specifically, we extract features from multiple layers of the discriminator and learn to match these intermediate representations from the real and the synthesized image. For ease of presentation, we denote the ith-layer feature extractor of discriminator Dk as D(i) (from input to the ith layer of Dk). The feature matching loss LFM(G,Dk) is then calculated

%Although the discriminators have an identical architecture, the one that operates at the coarsest scale has the largest receptive field. It has a more global view of the image and can guide the gener- ator to generate globally consistent images. On the other hand, the discriminator at the finest scale encourages the generator to produce finer details. This also makes training the coarse-to-fine generator easier, since extending a low- resolution model to a higher resolution only requires adding a discriminator at the finest level, rather than retraining from scratch. Without the multi-stage discriminators, we observe that many repeated patterns often appear in the generated images.

\subsection{Objective Functions}
To train the model, we follow the ControlGAN \cite{li2019controllable} and add extra unconditional adversarial losses ($\mathcal{L}_{Z_{Di}}$, $\mathcal{L}_{Z_{Gi}}$) in Eqs.\ref{eq:discriminator} and \ref{eq:generator} and the structure loss ($\mathcal{L}_{S_{i}}$) in Eq.~\ref{eq:SL} at each stage. Generators and discriminators are optimised alternatively by minimising their objective functions. Please see the supplementary material for complete objectives. We only the highlight differences compared to the ControlGAN.

%To train the network, we follow [14] and adopt adversarial training, where our network and the discriminators (D1 , D2 , D3 , DDCM ) are alternatively optimised. Please see supplementary material for more details about training objectives. We only highlight some training differences compared with [14].

% In the unusual situation where you want a paper to appear in the references without citing it in the main text, use \nocite
\section{Experiments}
The model is evaluated on the COCO dataset \cite{lin2014microsoft}, generating different-scale images progressively. We are unaware of any previous end-to-end methods for generating images from segmentation masks with embedded controllable factors, e.g., natural language descriptions, so we compare our method with AttnGAN \cite{xu2018attngan} and ControlGAN \cite{li2019controllable}, state-of-the-art methods for generating images from texts. To have a fair comparison, we slightly modify both models by implementing ACM \cite{li2019manigan} to incorporate segmentation masks, and call the modified models S-AttGAN and S-ControlGAN, respectively. 
%There are works \cite{johnson2018image, hong2018inferring, ashual2019specifying} that proposed to generate intermediate layout first, i.e., bounding boxes and/or segmentation masks, and then convert the semantic layout into an image. However, 
Note that we do not choose the models introduced in \cite{johnson2018image, ashual2019specifying} as baselines, because input scene graphs do not contain descriptions of visual attributes. Also, we do not compare our work with studies \cite{hong2018inferring, li2019object}, because models proposed in both studies require bounding boxes as intermediate input to produce synthetic images.

%the truth is synthetic images are still far from satisfactory with low-quality.
 
%In order to generate realistic images on COCO, \cite{johnson2018image, ashual2019specifying, hong2018inferring} proposed to generate intermediate layout first, i.e., bounding boxes and/or segmentation masks, and then convert the semantic layout into an image. 
%However, the truth is that the synthetic images are still far from satisfactory with low-resolution. Although these model break the complicated generation process into two more tractable stages, it is still a cross-domain translation, no matter the output is segmentation masks or realistic images. 

%, providing ground-truth segmentation masks as intermediate semantic input.

\textbf{Dataset.} The COCO \cite{lin2014microsoft} contains $82,783$ training images and $40,504$ validation images. Each image has a ground truth semantic mask and 5 descriptions. In our task, we only use binary segmentation masks instead of fine-grained pixel-labelled semantic maps. We preprocess the dataset according to the method in \cite{zhang2017stackgan}. 

\textbf{Implementation.} Our model has three stages and each stage has a generator and a discriminator. 
Three different-scale images ($64 \times 64$, $128 \times 128$, and $256 \times 256$) are generated progressively. %Discriminators at different stages have different network structures, i.e., the discriminator at the higher stage is deeper in order to have a larger receptive field. 
The model is trained for 120 epochs on the COCO dataset using the Adam optimiser \cite{kingma2014adam} with the learning rate $0.0002$. The hyperparameters controlling the extra losses $\mathcal{L}_{Z_{D}}$, $\mathcal{L}_{Z_{G}}$ and $\mathcal{L}_{S}$ are set to 1.
\noindent\begin{figure*}[t]
\begin{minipage}{0.092\textwidth}
\raggedright
\scriptsize{Text}
\end{minipage}
\begin{minipage}{0.105\textwidth}
\centering
\scriptsize{Orange tree with ripe \textbf{oranges} and \textbf{green leaves}.}
\end{minipage}
\begin{minipage}{0.105\textwidth}
\centering
\scriptsize{Several \textbf{donuts} are on a table.}
\end{minipage}
\;\begin{minipage}{0.105\textwidth}
\centering
\scriptsize{A \textbf{yellow} bus parks in the road.}
\end{minipage}
\begin{minipage}{0.105\textwidth}
\centering
\scriptsize{A \textbf{red} bus parks in the road.}
\end{minipage}
\;\begin{minipage}{0.105\textwidth}
\centering
\scriptsize{A large pizza with cheese and pepperoni on a white plate.}
\end{minipage}
\begin{minipage}{0.105\textwidth}
\centering
\scriptsize{A large pizza with cheese, pepperoni and \textbf{fresh herbs} on a white plate.}
\end{minipage}
\;\begin{minipage}{0.105\textwidth}
\centering
\scriptsize{A giraffe is walking on a dirt road under a \textbf{blue sky}.}
\end{minipage}
\begin{minipage}{0.105\textwidth}
\centering
\scriptsize{A giraffe is walking on a dirt road under a \textbf{sunset sky}.}
\end{minipage}

\noindent\begin{minipage}{0.092\textwidth}
\raggedright
\scriptsize{Segmentation Mask}
\end{minipage}
\noindent\begin{minipage}{0.105\textwidth}
\includegraphics[width=1\linewidth, height=1\linewidth]{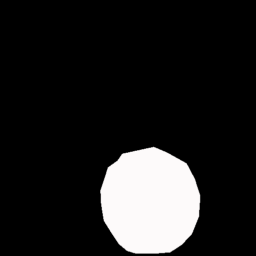}
\end{minipage}
\noindent\begin{minipage}{0.105\textwidth}
\includegraphics[width=1\linewidth, height=1\linewidth]{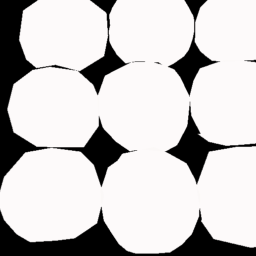}
\end{minipage}
\;\begin{minipage}{0.105\textwidth}
\includegraphics[width=1\linewidth, height=1\linewidth]{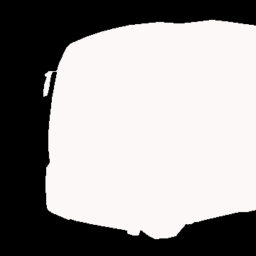}
\end{minipage}
\noindent\begin{minipage}{0.105\textwidth}
\includegraphics[width=1\linewidth, height=1\linewidth]{qual_3_seg.png}
\end{minipage}
\;\begin{minipage}{0.105\textwidth}
\includegraphics[width=1\linewidth, height=1\linewidth]{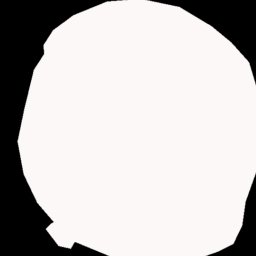}
\end{minipage}
\noindent\begin{minipage}{0.105\textwidth}
\includegraphics[width=1\linewidth, height=1\linewidth]{qual_4_seg.png}
\end{minipage}
\;\begin{minipage}{0.105\textwidth}
\includegraphics[width=1\linewidth, height=1\linewidth]{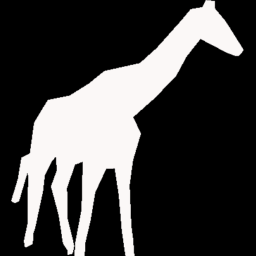}
\end{minipage}
\noindent\begin{minipage}{0.105\textwidth}
\includegraphics[width=1\linewidth, height=1\linewidth]{qual_1_seg.png}
\end{minipage}

\noindent\begin{minipage}{0.092\textwidth}
\raggedright
\scriptsize{S-AttnGAN}
\end{minipage}
\noindent\begin{minipage}{0.105\textwidth}
\includegraphics[width=1\linewidth, height=1\linewidth]{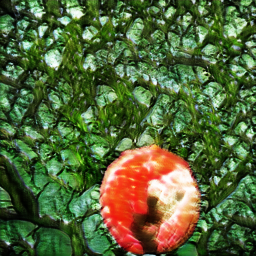}
\end{minipage}
\noindent\begin{minipage}{0.105\textwidth}
\includegraphics[width=1\linewidth, height=1\linewidth]{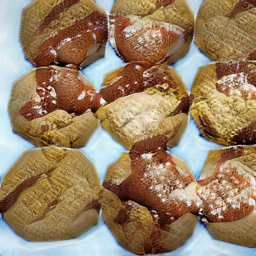}
\end{minipage}
\;\begin{minipage}{0.105\textwidth}
\includegraphics[width=1\linewidth, height=1\linewidth]{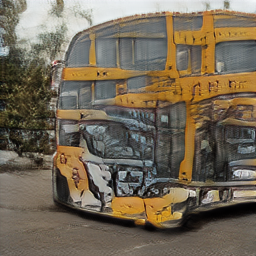}
\end{minipage}
\noindent\begin{minipage}{0.105\textwidth}
\includegraphics[width=1\linewidth, height=1\linewidth]{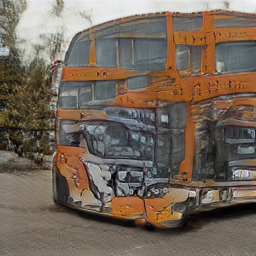}
\end{minipage}
\;\begin{minipage}{0.105\textwidth}
\includegraphics[width=1\linewidth, height=1\linewidth]{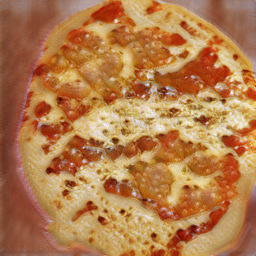}
\end{minipage}
\noindent\begin{minipage}{0.105\textwidth}
\includegraphics[width=1\linewidth, height=1\linewidth]{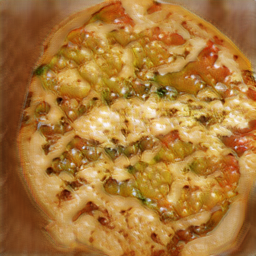}
\end{minipage}
\;\begin{minipage}{0.105\textwidth}
\includegraphics[width=1\linewidth, height=1\linewidth]{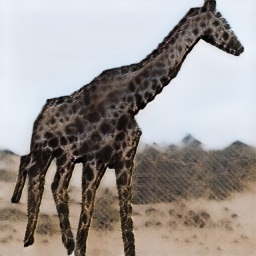}
\end{minipage}
\noindent\begin{minipage}{0.105\textwidth}
\includegraphics[width=1\linewidth, height=1\linewidth]{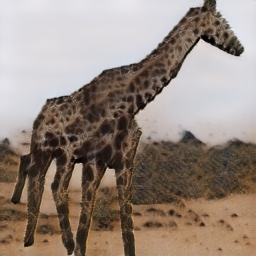}
\end{minipage}

\noindent\begin{minipage}{0.092\textwidth}
\raggedright
\scriptsize{S-ControlGAN}
\end{minipage}
\noindent\begin{minipage}{0.105\textwidth}
\includegraphics[width=1\linewidth, height=1\linewidth]{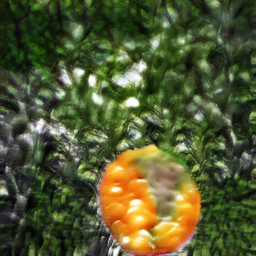}
\end{minipage}
\noindent\begin{minipage}{0.105\textwidth}
\includegraphics[width=1\linewidth, height=1\linewidth]{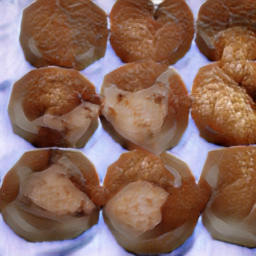}
\end{minipage}
\;\begin{minipage}{0.105\textwidth}
\includegraphics[width=1\linewidth, height=1\linewidth]{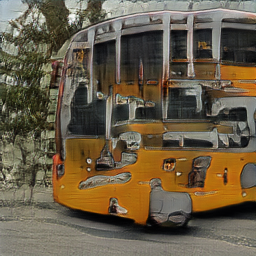}
\end{minipage}
\noindent\begin{minipage}{0.105\textwidth}
\includegraphics[width=1\linewidth, height=1\linewidth]{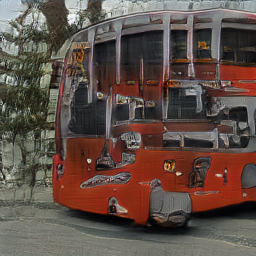}
\end{minipage}
\;\begin{minipage}{0.105\textwidth}
\includegraphics[width=1\linewidth, height=1\linewidth]{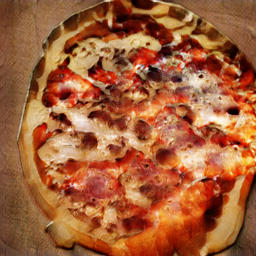}
\end{minipage}
\noindent\begin{minipage}{0.105\textwidth}
\includegraphics[width=1\linewidth, height=1\linewidth]{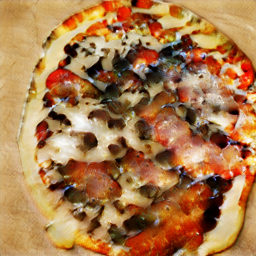}
\end{minipage}
\;\begin{minipage}{0.105\textwidth}
\includegraphics[width=1\linewidth, height=1\linewidth]{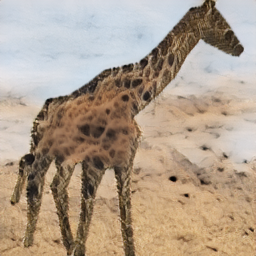}
\end{minipage}
\noindent\begin{minipage}{0.105\textwidth}
\includegraphics[width=1\linewidth, height=1\linewidth]{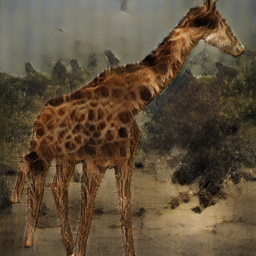}
\end{minipage}

\begin{minipage}{0.092\textwidth}
\raggedright
\scriptsize{Ours}
\end{minipage}
\begin{minipage}{0.105\textwidth}
\includegraphics[width=1\linewidth, height=1\linewidth]{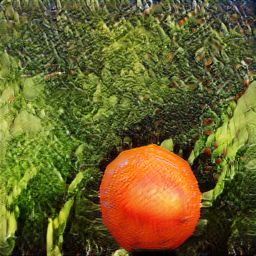}
\end{minipage}
\begin{minipage}{0.105\textwidth}
\includegraphics[width=1\linewidth, height=1\linewidth]{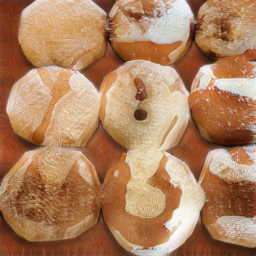}
\end{minipage}
\;\begin{minipage}{0.105\textwidth}
\includegraphics[width=1\linewidth, height=1\linewidth]{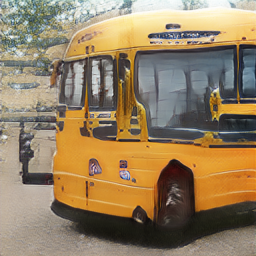}
\end{minipage}
\begin{minipage}{0.105\textwidth}
\includegraphics[width=1\linewidth, height=1\linewidth]{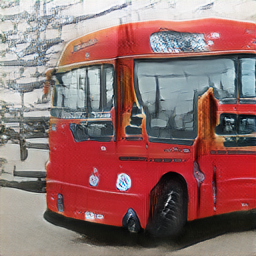}
\end{minipage}
\;\begin{minipage}{0.105\textwidth}
\includegraphics[width=1\linewidth, height=1\linewidth]{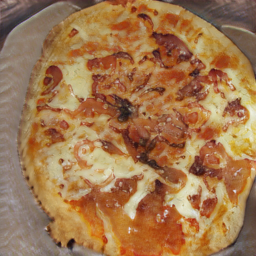}
\end{minipage}
\begin{minipage}{0.105\textwidth}
\includegraphics[width=1\linewidth, height=1\linewidth]{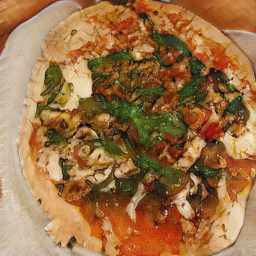}
\end{minipage}
\;\begin{minipage}{0.105\textwidth}
\includegraphics[width=1\linewidth, height=1\linewidth]{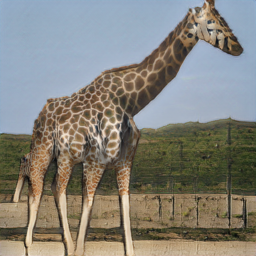}
\end{minipage}
\begin{minipage}{0.105\textwidth}
\includegraphics[width=1\linewidth, height=1\linewidth]{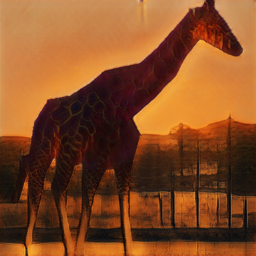}
\end{minipage}
\smallskip 

\begin{minipage}{0.092\textwidth}
\raggedright
\qquad\qquad\qquad\qquad\qquad
\end{minipage}
\begin{minipage}{0.105\textwidth}
\centering
\footnotesize{a}
\end{minipage}
\begin{minipage}{0.105\textwidth}
\centering
\footnotesize{b}
\end{minipage}
\;\begin{minipage}{0.105\textwidth}
\centering
\footnotesize{c}
\end{minipage}
\begin{minipage}{0.105\textwidth}
\centering
\footnotesize{d}
\end{minipage}
\;\begin{minipage}{0.105\textwidth}
\centering
\footnotesize{e}
\end{minipage}
\begin{minipage}{0.105\textwidth}
\centering
\footnotesize{f}
\end{minipage}
\;\begin{minipage}{0.105\textwidth}
\centering
\footnotesize{g}
\end{minipage}
\begin{minipage}{0.105\textwidth}
\centering
\footnotesize{h}
\end{minipage}

\centering
\caption{Qualitative comparison of three methods on the COCO dataset. (1) \emph{a} and \emph{b} represent the generation of objects belonging to different categories on similar segmentation masks; (2) \emph{c} and \emph{d} illustrate the controllable ability of internal visual attributes of objects; (3) \emph{e} and \emph{f} show the capability of adding new visual attributes on synthetic images while preserving other text-unmodified contents; and (4) \emph{g} and \emph{h} show that the model can also control the global style of the generated results.}
\label{fig:qual}
\end{figure*}
\subsection{Comparison with State-of-the-Art Approaches}
\begin{table}[h!]
  \centering
  \vspace{-5mm}
  \caption{Quantitative comparison: Inception Score (IS) and R-precision (R-prcn) of state-of-the-art methods and RefinedGAN on the COCO dataset. ``w/o POS" denotes without part-of-speech tagging; ``w/ Concate." denotes using a concatenation method to combine text and image features instead of using the affine combination module; ``w/o Refined" denotes without using refined multi-stage architectures on discriminators and generators; ``w/o SL" denotes without structure loss; ``w/o POS*" denotes the model is trained without using POS, but the POS is implemented on the testing. For IS and R-prcn, higher is better.}
  \smallskip
  \scalebox{0.90}{
  \begin{tabular}{|l||cc|}
    \hline
    Method & IS & R-prcn (\%) \\
    \hline
    \hline
    Real Images & 27.41 $\pm$ 0.59 & - \\
    \hline
    S-AttnGAN  & 12.09 $\pm$ 0.28 & 75.24 $\pm$ 3.39   \\
    S-ControlGAN  & 11.56 $\pm$ 0.16 & 80.43 $\pm$ 2.79  \\
    \hline
    Ours w/o POS  &  \textbf{16.49 $\pm$ 0.18} & \textbf{84.01 $\pm$ 1.59} \\
    Ours w/ Concat.   & 8.50 $\pm$ 0.15 & 44.11 $\pm$ 3.99  \\
    Ours w/o Refined   & 12.16 $\pm$ 0.20 & 80.13 $\pm$ 2.20  \\
    Ours w/o SL   & 14.72 $\pm$ 0.32 & 81.43 $\pm$ 1.21  \\
    \hline
    Ours w/o POS*  & 14.74 $\pm$ 0.13 & 83.03 $\pm$ 1.15  \\
    \hline
    \textbf{Ours}   & 15.96 $\pm$ 0.16 & 83.23 $\pm$ 1.37  \\
    \hline
  \end{tabular}
  }
\label{table:quantitative}
\end{table}

\textbf{Quantitative comparison.}
We adopt the Inception Score (IS) \cite{xu2018attngan} to evaluate the quality and diversity of synthetic images. 
%, and the \text{Fr$\acute{e}$chet} Inception Distance (FID) \cite{heusel2017gans} to measure the distance between the distribution of synthetic images and that of the real images.
Also, to measure the semantic consistency between the generated images and the corresponding text descriptions, we adopt the R-precision (R-prcn) \cite{xu2018attngan}, which is an evaluation metric for ranking retrieval results. IS and R-prcn are evaluated on a large number of matched text-mask pairs sampled from the dataset.
%synthetic and original images cropped by the given segmentation mask. 
%In our experiments, IS and R-prcn are evaluated on a large number of matched pairs, i.e., text descriptions correctly matching the corresponding segmentation masks. As for Diff, the modified texts are randomly sampled from dataset belonging to the same class of the original texts and corresponding masks. 
%In order to have fair comparison, the ground truth semantic layout (i.e., segmentation mask) is provided on both S2IM \cite{johnson2018image} and SG \cite{ashual2019specifying} models. Also, we slightly modify the code of S2IM provided by authors to produce images with $256 \times 256$ scale. In our experiments, IS, FID, SIM and DIFF are all evaluated on a large number of pair samples, i.e., text descriptions correctly matching corresponding segmentation masks. %, and  is evaluated on a large number of mismatched pairs, i.e., randomly chosen segmentation maps with randomly selected text descriptions. 

\begin{figure}[t]
\begin{minipage}{0.157\textwidth}
\centering
\quad
\end{minipage}
\begin{minipage}{0.157\textwidth}
\centering
\scriptsize{A blue bus that is outside of a building.}
\end{minipage}
\begin{minipage}{0.157\textwidth}
\centering
\scriptsize{Blue bus outside of building.}
\end{minipage}

\noindent\begin{minipage}{0.157\textwidth}
\includegraphics[width=1\linewidth, height=1\linewidth]{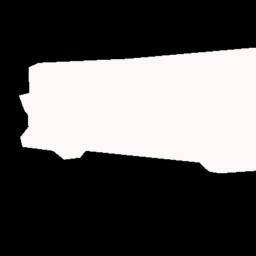}
\end{minipage}
\noindent\begin{minipage}{0.157\textwidth}
\includegraphics[width=1\linewidth, height=1\linewidth]{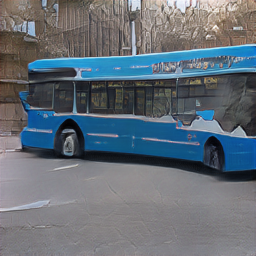}
\end{minipage}
\begin{minipage}{0.157\textwidth}
\includegraphics[width=1\linewidth, height=1\linewidth]{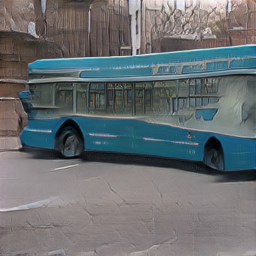}
\end{minipage}
\smallskip

\begin{minipage}{0.157\textwidth}
\centering
\quad
\end{minipage}
\begin{minipage}{0.157\textwidth}
\centering
\scriptsize{A couple of kites are flying in the blue sky.}
\end{minipage}
\begin{minipage}{0.157\textwidth}
\centering
\scriptsize{Kites flying in blue sky.}
\end{minipage}

\noindent\begin{minipage}{0.157\textwidth}
\includegraphics[width=1\linewidth, height=1\linewidth]{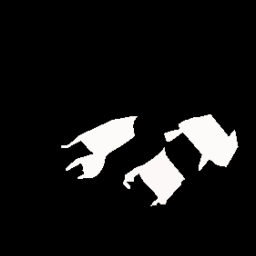}
\end{minipage}
\noindent\begin{minipage}{0.157\textwidth}
\includegraphics[width=1\linewidth, height=1\linewidth]{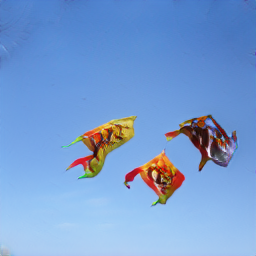}
\end{minipage}
\begin{minipage}{0.157\textwidth}
\includegraphics[width=1\linewidth, height=1\linewidth]{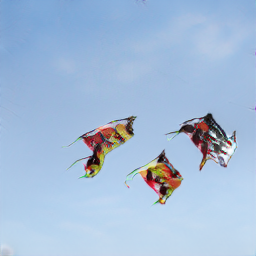}
\end{minipage}
\smallskip

\begin{minipage}{0.157\textwidth}
\centering
\scriptsize{a: Segmentation mask.}
\end{minipage}
\begin{minipage}{0.157\textwidth}
\centering
\scriptsize{b: Original text.}
\end{minipage}
\begin{minipage}{0.157\textwidth}
\centering
\scriptsize{c: Text without non-semantic words.}
\end{minipage}

\centering
\caption{Existence of unnecessary connections between non-semantic words and visual attributes. \emph{b} and \emph{c} show images that are generated from full text descriptions and filtered descriptions by the model without POS, respectively.}
\label{fig:exist}
\end{figure}

As shown in Table.~\ref{table:quantitative}, our model achieves a better IS value, which demonstrates that our model can generate more realistic images with high diversity. Also, the better R-prcn value indicates that the synthetic images generated by our model highly match the given text descriptions. Note that compared to ``Ours w/o POS", the IS and R-prcn values of ``Ours w/o POS*" decrease, which illustrates that unnecessary connections are built between non-semantic words and visual attributes, and these useless bonding can harm the quality of synthetic results, examples shown in Fig.~\ref{fig:exist}.
% and the generated objects are more identical to original ones.
\noindent\begin{figure}[t]
\begin{minipage}{0.232\textwidth}
\centering
\scriptsize{Segmentation mask.}
\end{minipage}
\;\begin{minipage}{0.232\textwidth}
\centering
\scriptsize{Giraffe standing around in the middle of a field with trees in the background.}
\end{minipage}

\noindent\begin{minipage}{0.116\textwidth}
\includegraphics[width=1\linewidth, height=1\linewidth]{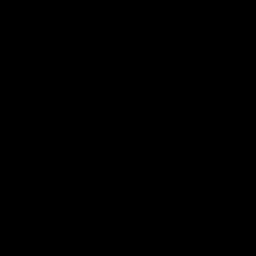}
\end{minipage}
\noindent\begin{minipage}{0.116\textwidth}
\includegraphics[width=1\linewidth, height=1\linewidth]{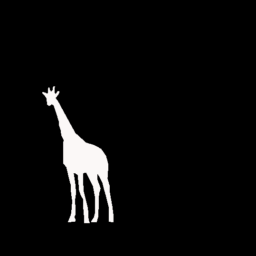}
\end{minipage}
\;\begin{minipage}{0.116\textwidth}
\includegraphics[width=1\linewidth, height=1\linewidth]{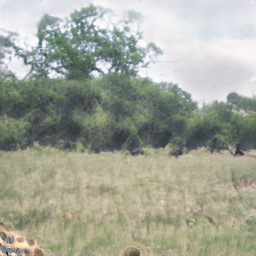}
\end{minipage}
\noindent\begin{minipage}{0.116\textwidth}
\includegraphics[width=1\linewidth, height=1\linewidth]{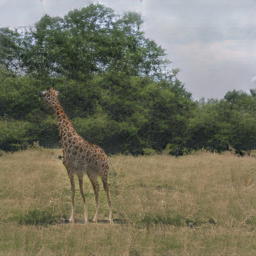}
\end{minipage}

\noindent\begin{minipage}{0.116\textwidth}
\includegraphics[width=1\linewidth, height=1\linewidth]{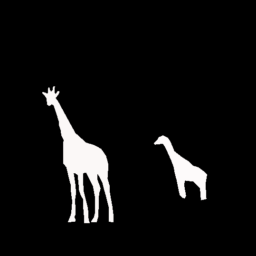}
\end{minipage}
\noindent\begin{minipage}{0.116\textwidth}
\includegraphics[width=1\linewidth, height=1\linewidth]{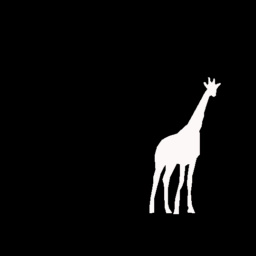}
\end{minipage}
\;\begin{minipage}{0.116\textwidth}
\includegraphics[width=1\linewidth, height=1\linewidth]{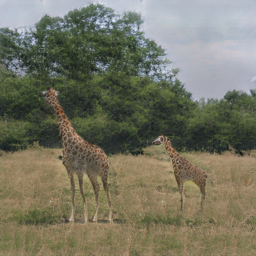}
\end{minipage}
\noindent\begin{minipage}{0.116\textwidth}
\includegraphics[width=1\linewidth, height=1\linewidth]{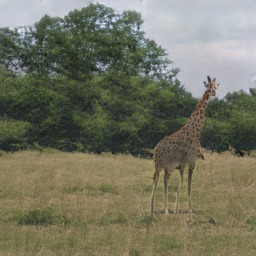}
\end{minipage}

\centering
\caption{Disentanglement of objects and background.}
\label{fig:disentangle}
\end{figure}
\begin{figure*}[t]
\begin{minipage}{0.158\textwidth}
\centering
\scriptsize{A \textbf{white} bus is pulling up to a bus stop in a city.}
\end{minipage}
\;\begin{minipage}{0.163\textwidth}
\includegraphics[width=1\linewidth, height=1\linewidth]{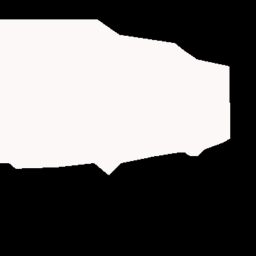}
\end{minipage}
\begin{minipage}{0.163\textwidth}
\includegraphics[width=1\linewidth, height=1\linewidth]{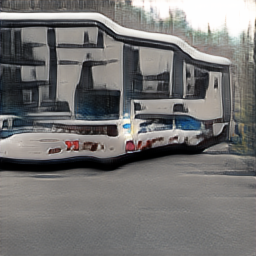}
\end{minipage}
\begin{minipage}{0.163\textwidth}
\includegraphics[width=1\linewidth, height=1\linewidth]{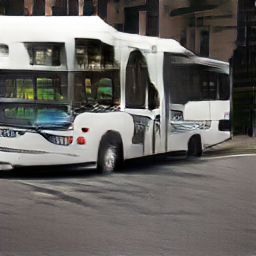}
\end{minipage}
\begin{minipage}{0.163\textwidth}
\includegraphics[width=1\linewidth, height=1\linewidth]{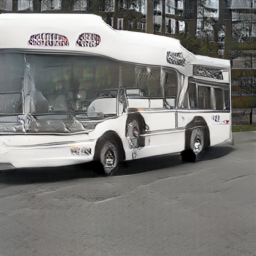}
\end{minipage}
\begin{minipage}{0.163\textwidth}
\includegraphics[width=1\linewidth, height=1\linewidth]{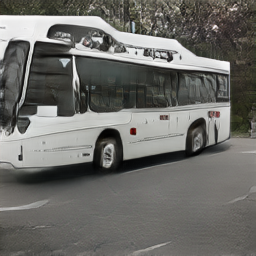}
\end{minipage}

\begin{minipage}{0.158\textwidth}
\centering
\scriptsize{A \textbf{red} commerical airplane is flying over a \textbf{clear sky}.}
\end{minipage}
\;\begin{minipage}{0.163\textwidth}
\includegraphics[width=1\linewidth, height=1\linewidth]{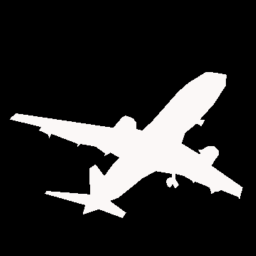}
\end{minipage}
\begin{minipage}{0.163\textwidth}
\includegraphics[width=1\linewidth, height=1\linewidth]{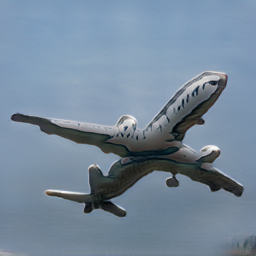}
\end{minipage}
\begin{minipage}{0.163\textwidth}
\includegraphics[width=1\linewidth, height=1\linewidth]{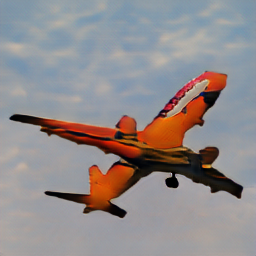}
\end{minipage}
\begin{minipage}{0.163\textwidth}
\includegraphics[width=1\linewidth, height=1\linewidth]{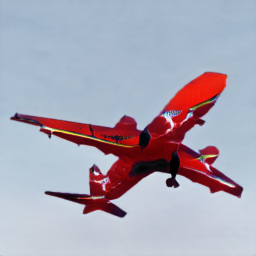}
\end{minipage}
\begin{minipage}{0.163\textwidth}
\includegraphics[width=1\linewidth, height=1\linewidth]{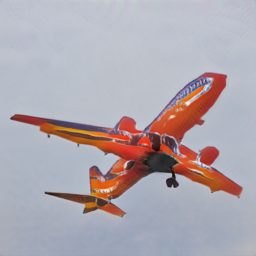}
\end{minipage}
\smallskip

\begin{minipage}{0.158\textwidth}
\centering
\scriptsize{a: Text}
\end{minipage}
\;\begin{minipage}{0.163\textwidth}
\centering
\scriptsize{b: Segmentation mask}
\end{minipage}
\begin{minipage}{0.163\textwidth}
\centering
\scriptsize{c: Our w/ Concat.}
\end{minipage}
\begin{minipage}{0.163\textwidth}
\centering
\scriptsize{d: Our w/o Refined}
\end{minipage}
\begin{minipage}{0.163\textwidth}
\centering
\scriptsize{e: Our w/o SL}
\end{minipage}
\begin{minipage}{0.163\textwidth}
\centering
\scriptsize{f: Ours}
\end{minipage}

\centering
\caption{Ablation studies. \emph{a}: given text description with the desired objects and visual attributes; \emph{b}: segmentation mask; \emph{c}: using the concatenation method to replace the affine combination module; \emph{d}: without implementing the refined multi-stage architecture; \emph{e}: without structure loss; \emph{f}: our full model.}
\label{fig:ablation}
\end{figure*}
\noindent

\textbf{Qualitative comparison.}
Fig.~\ref{fig:qual} shows the visual comparison between the RefinedGAN, AttnGAN-Seg, and ControlGAN-Seg on the COCO dataset \cite{lin2014microsoft}.
%: (1) \emph{a} and \emph{b} represent the generation of objects belonging different categories; (2) \emph{c} and \emph{d} illustrate the controllable ability of internal visual attributes; (3) \emph{e} and \emph{d} show the capability of adding new visual attributes on synthetic images; and (4) \emph{g} and \emph{h} show that the model can determine the global style of results.
We can easily observe that both methods are only able to generate low-quality images with unrealistic objects, e.g., the buses produced by both methods have distorted textures (columns \emph{c} and \emph{d}), and oranges have unusual black or brown colour (column \emph{a}). Also, the synthetic results generated from both methods do not have a perfect semantic consistency with given text descriptions, i.e., both methods fail to produce the new global style \emph{sunset} at the column \emph{h}, and the new attribute \emph{fresh herbs} at the column \emph{f}. %On the contrary, our model can generate realistic images semantically matching the given descriptions. 

Failing to generate realistic images by both methods is mainly because: (1) there exist unnecessary connections between non-semantic words and visual attributes, which can potentially constrain the control of attributes (e.g., change of colour to \emph{red} at \emph{d}); and (2) both methods fail to produce a complete structure at lower stages and also do not have an effective rectification ability, e.g., no \emph{white plate} at the column \emph{e}, no \emph{blue sky} at column \emph{g}. However, our model addresses above problems by implementing POS tagging, the refined multi-stage architecture and structure loss. More details are discussed in Sec.\ref{sec:ablation}.

Besides, our model can effectively disentangle the foreground objects with background, shown in Fig.~\ref{fig:disentangle}. As we can see that if there is no segmentation mask being provided, only background is generated by our model, but the result still semantically matches the given description. Also, the generation of objects has almost no impact on the generation of background, even when we provide different segmentation masks, which illustrates an effective disentanglement between foreground and background. Based on this, our model can generate diverse results by adding objects without changing the background, and also enable us to modify visual attributes of synthetic images, while preserving content that is not required in the modified text descriptions. For example, shown in the columns \emph{e} and \emph{f} in Fig.~\ref{fig:qual}, the backgrounds \emph{white plate} are almost the same when a new attribute \emph{fresh herbs} is added.
%it is really easy to add objects on an image by providing corresponding segmentation masks.
%chosen by users. To achieve this, they can simply crop objects from one generated images and merge them on the other image.
%Furthermore, the model can generate objects sequentially by providing the segmentation masks one by one, shown in the Fig.XX. Also, we can observe that our model is able to add required objects while preserving the background similar as previous one, which illustrates effective disentanglement between foreground and background. Based on this, our model can generate diverse results by adding objects without changing other contents, and also enable to modify visual attributes of synthetic images while preserving contents that are not required in the modified text description.
\begin{figure*}[t]
\begin{minipage}{0.120\textwidth}
\centering
\scriptsize{A \textbf{zebra} is grazing on \textbf{green grass}.}
\end{minipage}
\begin{minipage}{0.120\textwidth}
\includegraphics[width=1\linewidth, height=1\linewidth]{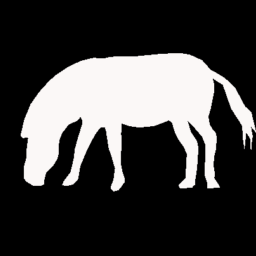}
\end{minipage}
\;\begin{minipage}{0.120\textwidth}
\includegraphics[width=1\linewidth, height=1\linewidth]{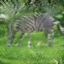}
\end{minipage}
\begin{minipage}{0.120\textwidth}
\includegraphics[width=1\linewidth, height=1\linewidth]{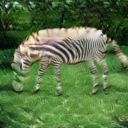}
\end{minipage}
\begin{minipage}{0.120\textwidth}
\includegraphics[width=1\linewidth, height=1\linewidth]{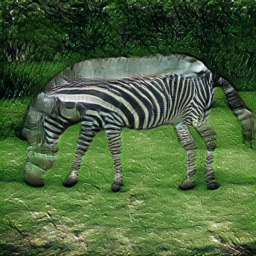}
\end{minipage}
\;\begin{minipage}{0.120\textwidth}
\includegraphics[width=1\linewidth, height=1\linewidth]{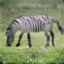}
\end{minipage}
\begin{minipage}{0.120\textwidth}
\includegraphics[width=1\linewidth, height=1\linewidth]{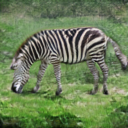}
\end{minipage}
\begin{minipage}{0.120\textwidth}
\includegraphics[width=1\linewidth, height=1\linewidth]{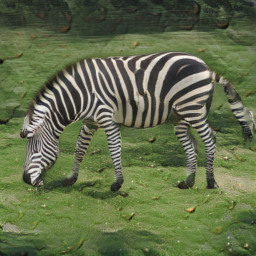}
\end{minipage}
\smallskip

\begin{minipage}{0.120\textwidth}
\centering
\scriptsize{A large \textbf{giraffe} and a small \textbf{giraffe} are standing in front of \textbf{trees}.}
\end{minipage}
\begin{minipage}{0.120\textwidth}
\includegraphics[width=1\linewidth, height=1\linewidth]{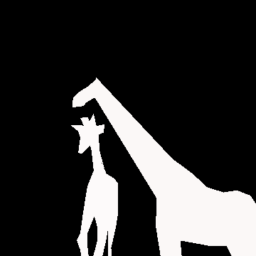}
\end{minipage}
\;\begin{minipage}{0.120\textwidth}
\includegraphics[width=1\linewidth, height=1\linewidth]{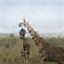}
\end{minipage}
\begin{minipage}{0.120\textwidth}
\includegraphics[width=1\linewidth, height=1\linewidth]{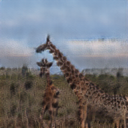}
\end{minipage}
\begin{minipage}{0.120\textwidth}
\includegraphics[width=1\linewidth, height=1\linewidth]{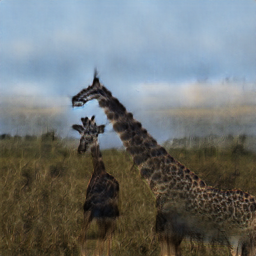}
\end{minipage}
\;\begin{minipage}{0.120\textwidth}
\includegraphics[width=1\linewidth, height=1\linewidth]{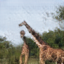}
\end{minipage}
\begin{minipage}{0.120\textwidth}
\includegraphics[width=1\linewidth, height=1\linewidth]{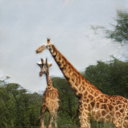}
\end{minipage}
\begin{minipage}{0.120\textwidth}
\includegraphics[width=1\linewidth, height=1\linewidth]{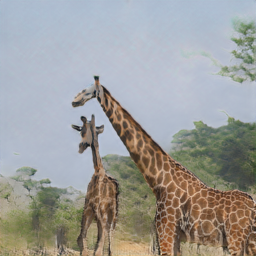}
\end{minipage}
\smallskip

\begin{minipage}{0.120\textwidth}
\centering
\scriptsize{a: Text}
\end{minipage}
\begin{minipage}{0.120\textwidth}
\centering
\scriptsize{b: Segmentation mask}
\end{minipage}
\;\begin{minipage}{0.120\textwidth}
\centering
\scriptsize{c: Stage 1, w/o Refined}
\end{minipage}
\begin{minipage}{0.120\textwidth}
\centering
\scriptsize{d: Stage 2, w/o Refined}
\end{minipage}
\begin{minipage}{0.120\textwidth}
\centering
\scriptsize{e: Stage 3, w/o Refined}
\end{minipage}
\;\begin{minipage}{0.120\textwidth}
\centering
\scriptsize{f: Stage 1, Our}
\end{minipage}
\begin{minipage}{0.120\textwidth}
\centering
\scriptsize{g: Stage 2, Our}
\end{minipage}
\begin{minipage}{0.120\textwidth}
\centering
\scriptsize{h: Stage 3, Our}
\end{minipage}

\centering
\caption{Effectiveness of refined multi-stage architecture. \emph{c}, \emph{d}, and \emph{e} show the synthetic images produced at each stage by the model without refined multi-stage architecture. \emph{f}, \emph{g}, and \emph{h} show the synthetic images generated at each stage by our model.}
\label{fig:refined}
\end{figure*}
\noindent\begin{figure}[t]
\noindent\begin{minipage}{0.157\textwidth}
\includegraphics[width=1\linewidth, height=1\linewidth]{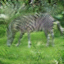}
\end{minipage}
\noindent\begin{minipage}{0.157\textwidth}
\includegraphics[width=1\linewidth, height=1\linewidth]{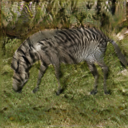}
\end{minipage}
\begin{minipage}{0.157\textwidth}
\includegraphics[width=1\linewidth, height=1\linewidth]{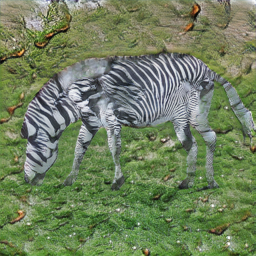}
\end{minipage}
\smallskip

\noindent\begin{minipage}{0.157\textwidth}
\includegraphics[width=1\linewidth, height=1\linewidth]{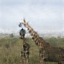}
\end{minipage}
\noindent\begin{minipage}{0.157\textwidth}
\includegraphics[width=1\linewidth, height=1\linewidth]{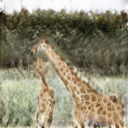}
\end{minipage}
\begin{minipage}{0.157\textwidth}
\includegraphics[width=1\linewidth, height=1\linewidth]{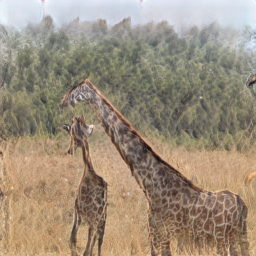}
\end{minipage}
\smallskip

\begin{minipage}{0.157\textwidth}
\centering
\scriptsize{a: Stage 1, w/o Refined}
\end{minipage}
\begin{minipage}{0.157\textwidth}
\centering
\scriptsize{b: Stage 2, Our}
\end{minipage}
\begin{minipage}{0.157\textwidth}
\centering
\scriptsize{c: Stage 3, Our}
\end{minipage}

\centering
\caption{Rectification ability of generators in RefinedGAN. \emph{a} denotes images generated at the first stage by the model without refined architecture. \emph{b} and \emph{c} denote our model takes this flawed features and feed through stages 2 and 3 progressively, producing the corresponding images shown at \emph{b} and \emph{c}.}
\label{fig:rectification}
\end{figure}
\subsection{Ablation Studies}
\label{sec:ablation}
\textbf{Necessity of part-of-speech tagging.}
As discussed in Sec.~\ref{sec:pos}, the implementation of part-of-speech (POS) tagging can help to filter out specific words, especially less important ones, which can effectively prevent less useful information being contained in word and sentence features, and also avoid building inappropriate connections between non-semantic words and visual attributes, such that the model can achieve a better controllable performance. 

Thus, we conduct a study to verify the existence of those useless or even harmful connections via the model without POS, shown in Fig.~\ref{fig:exist}. As we can see, when we remove non-semantic words from the description, some regions of the synthetic image become unrealistic, i.e., there is an obvious white patch shown on the bus.
%the quality of synthetic image has changed significantly (see top), 
Moreover, those useless words can even decrease the quality of synthetic results, shown in the bottom of Fig.~\ref{fig:exist}, which reduces the brightness of the blue sky and affects the texture of colourful kites.
%fail to generate the XX. Also, the ablation study shown in Fig.~\ref{fig:ablation} verify the negative impact of those redundant and less useful words.

\textbf{Effectiveness of affine combination module.}
As shown in Fig.~\ref{fig:ablation} \emph{c} and \emph{f}, we conduct an ablation study to show the effectiveness of the affine combination module (ACM). As we can see, the adoption of concatenation instead of ACM to combine different modality features, the model cannot produce realistic images with finer details (see top of \emph{c}), and even fails to keep a semantic consistency with the given description (see bottom of \emph{c}), while our model is able to produce high-quality results with fine-grained visual attributes under the control of the texts. This is mainly because the simple concatenation cannot build an accurate connection between semantic words with corresponding regions of the image, and also fails to effectively encode the controllable text description into the generation process.

\textbf{Effectiveness of refined multi-stage architecture.}
To demonstrate the effectiveness of refined multi-stage architecture, an ablation study is conducted, shown in Fig.~\ref{fig:refined}. The model without the refined multi-stage architecture, i.e., without feeding patches of images at higher stages to lower ones, fails to produce completed images with appropriate regional details at the lower stages, especially the first stage, e.g., \emph{the zebra misses the back and head} at the top of columns \emph{c} and \emph{d}, and \emph{there is no tree background}, and \emph{the smaller giraffe misses legs} at the bottom of columns \emph{c} and \emph{d}.

Besides, the generators at the following stages fail to complete the missing content or rectify inappropriate attributes, and just leave it without any correction (see columns \emph{d} and \emph{e} of Fig.~\ref{fig:refined}). To further verify the rectification ability of the refined multi-stage architecture, we feed the flawed features generated by the model without refined multi-stage architecture at the first stage to our full model. As we can see in Fig.~\ref{fig:rectification}, even if there are missing parts in the given images, the generators in our full model are able to complete the missing attributes, e.g., \emph{adding back and head for the zebra} at the top row, and to correct inappropriate visual attributes, e.g., \emph{change the background with trees} at the bottom row.% with appropriate and fine-grained details.
%Compared with two columns x and xx, we can observe that the model without refined multi-stage architecture fails to get

Also, more examples in Fig.~\ref{fig:ablation} \emph{d} show that images produced by the model without refined multi-stage architecture keeps flawed and incorrect details without rectification, e.g., the inappropriate green window of the bus, and there is a stripe of texture missing on the head of airplane at bottom of \emph{d}.

\textbf{Effectiveness of structure loss.}
To verify the effectiveness of structure loss, we remove it from our model, shown in Fig.~\ref{fig:ablation} \emph{e}. As we can see, the synthetic images produced by the model without the structure loss contain some unrealistic regions, e.g., there are some unrealistic patches on the bus, and the appearance of the airplane is far from satisfactory. % but the background ``clear sky" is good. 
However, our model can generate not only realistic objects but also a high-quality background, which potentially demonstrates that the structure loss can improve the differential ability of discriminators to identify fake images if there exist small unreasonable areas only in objects or the background, and in turn improve generators to produce higher-quality images with finer details everywhere.

\section{Conclusion}
We have proposed a novel generative adversarial network, called RefinedGAN, which effectively embeds controllable factors, i.e., natural language descriptions, into image-to-image translation to control the generation of objects and visual attributes. Also, our model can disentangle objects from the background, produce complete images at lower stages and enable a great rectification ability.
Extensive experimental results demonstrate the advantages of our method, with respective to both high-quality image generation and the effectiveness of control of local visual attributes.

\newpage
\bibliography{example_paper}
\bibliographystyle{icml2020}


\end{document}